\pdfoutput=1
\documentclass[11pt]{article}

\usepackage[final]{acl}

\usepackage{times}
\usepackage{latexsym}

\usepackage[T1]{fontenc}
\usepackage[utf8]{inputenc}

\usepackage{microtype}

\usepackage{inconsolata}

\usepackage{graphicx}

\title{ POLygraph: Polish Fake News Dataset }

\author{ Daniel Dzienisiewicz, Filip Graliński, Piotr Jabłoński \\ {\bf Marek Kubis, Paweł Skórzewski, Piotr Wierzchoń } \\
        Adam Mickiewicz University, Poznań \\ ul. Wieniawskiego 1, 61-712 Poznań, Poland \\
        \texttt{\{dzienis,filip.gralinski,piotr.jablonski,} \\
        \texttt{marek.kubis,pawel.skorzewski,piotr.wierzchon\}@amu.edu.pl}}

\begin{document}
\maketitle

\begin{abstract}
This paper presents the POLygraph dataset, a unique resource for fake news detection in Polish. The dataset, created by an interdisciplinary team, is composed of two parts: the “fake-or-not” dataset with 11,360 pairs of news articles (identified by their URLs) and corresponding labels, and the “fake-they-say” dataset with 5,082 news articles (identified by their URLs) and tweets commenting on them. Unlike existing datasets, POLygraph encompasses a variety of approaches from source literature, providing a comprehensive resource for fake news detection. The data was collected through manual annotation by expert and non-expert annotators. The project also developed a software tool that uses advanced machine learning techniques to analyze the data and determine content authenticity. The tool and dataset are expected to benefit various entities, from public sector institutions to publishers and fact-checking organizations. Further dataset exploration will foster fake news detection and potentially stimulate the implementation of similar models in other languages.
The paper focuses on the creation and composition of the dataset, so it does not include a detailed evaluation of the software tool for content authenticity analysis, which is planned at a later stage of the project.
\end{abstract}

\section{Introduction}

This paper describes a dataset created for a project aimed at detecting and analyzing fake news on the Polish web. Fake news poses a significant threat in real-world situations, eroding trust in institutions, manipulating public opinion, and fueling societal tensions. To address this challenge, our project employs a unique hybrid research approach, merging narratological, comparative, and sociological techniques with natural language processing and big data analytics. An interdisciplinary team of experts in various fields, including mathematics, computer science, philology, media studies, law, philosophy, folklore, and IT, collaborates on this endeavor.
The project aims to develop a fake news detection software tool that uses a comprehensive database of sources, authors, and content, as well as advanced machine learning techniques and implicit trust ranking analyses to determine the authenticity of the content.

The dataset described in this paper consists of two parts. The first part, referred to as the “fake-or-not” dataset, contains 11,360 pairs of news articles (identified by URLs) and labels indicating whether the news is fake or not. The second part, known as the “fake-they-say” dataset, comprises 5,082 news articles (identified by URLs) and tweets commenting on them. Each tweet is accompanied by a label expressing the commentator’s opinion about the article's truthfulness.

Our software tool and its underlying dataset are intended to serve various beneficiaries, including public sector entities like the Ministry of Internal Affairs and Administration, the Ministry of Defense, the Police, the Internal Security Agency, and the Internal Security Service for public safety purposes. It could also be helpful for publishers, the Warsaw Stock Exchange, the Financial Supervision Commission (to monitor potential manipulations affecting company valuations or the country's macroeconomic status), fact-checking organizations, and analytical firms.

\section{Related Work} \label{sec:related-work}

\subsection{Tasks and Datasets}

In today's digital age, the rapid dissemination of information has led to an intertwined web of factual narratives and misinformation. The challenge of distinguishing between the two has spurred extensive research in various domains. Tasks such as fact verification \citep{schuster-etal-2019-towards,lewis20}, fact-checking \citep{wang-2017-liar,bhattarai-etal-2022-explainable}, fact-based text editing \citep{iso-etal-2020-fact}, and table-based fact verification \citep{Chen2020TabFact,eisenschlos-etal-2020-understanding} are crucial in this endeavor. The complexity is further heightened by the introduction of counterfactual elements, which encompass counterfactual detection \citep{yang-etal-2020-semeval}, inference \citep{pawlowski20,poulos21}, and explanation \citep{plumb20,ramon20}. Moreover, the classification of comments \citep{bornheim-etal-2021-fhac} based on their toxicity, engagement, and fact-claiming nature is an emerging area of interest.

The broader challenge of misinformation \citep{thorne-vlachos-2021-evidence,bhattarai-etal-2022-explainable} encapsulates various facets, including fake news detection \citep{shu17,wang-2017-liar}, deepfake detection \citep{rossler19,li20celebdf}, and fake image detection \citep{afchar18,rossler19}. The political sphere, as evidenced by stance detection tasks \citep{hanselowski-etal-2018-retrospective,borges19} related to the US 2020 Election \citep{kawintiranon-singh-2021-knowledge}, is particularly susceptible to these challenges. Complementary research areas such as hate speech detection \citep{davidson17,mathew21}, propaganda technique identification \citep{blaschke-etal-2020-cyberwalle}, aggression identification \citep{orasan-2018-aggressive,risch-krestel-2018-aggression}, satire detection \citep{li-etal-2020-multi-modal,ionescu21}, humor detection \citep{castro16,weller-seppi-2019-humor}, rumor detection \citep{kochkina-etal-2017-turing,zubiaga18,gorrell-etal-2019-semeval}, and deception detection \citep{guo23} further underscore the multifaceted nature of this challenge.

Several datasets and competitions, such as those hosted on Kaggle\footnote{\url{https://www.kaggle.com/c/fake-news}} and the ISOT Fake News Dataset \citep{ahmed2017detection,ahmed2018detecting}, have been developed to foster advancements in this domain. RumourEval competition \citep{gorrell-etal-2019-semeval} provided a dataset of dubious posts and ensuing conversations in social media, annotated both for stance and veracity. The competition received many submissions that used state-of-the-art methodology to tackle the challenges involved in rumor verification. Another example is the FEVER (Fact Extraction and VERification) dataset \citep{thorne-etal-2018-fever}, consisting of 185,445 claims generated by altering sentences from Wikipedia and subsequently classified without knowledge of the sentence they were derived from as “supported”, “refuted”, or “not enough info”.

For a comprehensive approach, it is imperative to integrate diverse sources, including fact-checking websites, encyclopedias, urban legends, conspiracy theories, and Wikipedia entries on fake news. Archival resources, such as the urban legend archive curated by \citet{gralinski2012znikajkaca}, offer unique insights. Furthermore, domain-specific datasets, focusing on works of sci-fi authors like Lem, Pratchett, and Sapkowski, or niche forums like Wykop.pl\footnote{\url{https://wykop.pl}} and Hyperreal\footnote{\url{https://hyperreal.info}}, provide a rich tapestry of data for analysis. An example of such a dataset is BAN-PL \citep{kolos-etal-2024-ban-pl}, collecting content from the Wykop.pl web service that contains offensive language, which makes an essential contribution to the automated detection of such language online, including hate speech and cyberbullying.

Our methodology for categorizing fake news and non-fake news is anchored in established guidelines, as outlined by resources like EUfactcheck\footnote{\url{https://eufactcheck.eu/wp-content/uploads/2020/02/EUfactcheck-manual-DEF2.pdf}}. Additionally, the emergence of fake news detectors, evident in browser plugins and extensions such as SurfSafe\footnote{\url{https://www.getsurfsafe.com/}}, Reality Defender\footnote{\url{https://realitydefender.com}}, or Fake News Chrome Extension\footnote{\url{https://tlkh.github.io/fake-news-chrome-extension}}, presents promising avenues for real-time misinformation mitigation.

This research aims to introduce a comprehensive Polish fake news dataset to lay a robust foundation for future endeavors in the realm of misinformation detection and analysis within the Polish context.

\subsection{Annotation Methodologies}

The current fake news detection techniques can be classified into several groups. For instance, according to \citet{wangetal}, there are three categories of methods: propagation structure-based, user information-based, and news content-based. Propagation structure-based methods involve extracting features related to news dissemination in social media. User information-based methods focus on the users involved in the circulation of news, covering aspects such as users’ gender, social media friends, followers, and location. On the other hand, news content-based methods concentrate solely on analyzing the content of the news rather than information about users and news dissemination.

A mixed approach to fake news detection was proposed by \citet{zhangghorbani}, who identified four components considered particularly important in characterizing fake news: creator/disseminator, target, news content, and social context.
\citet{zhouzafarani}, on the other hand, divide fake news detection models into methods based on the analysis of the annotator’s knowledge (knowledge-based fake news detection), the style in which the news is written (style-based fake news detection), the method of disseminating the news (propagation-based fake news detection), and assessing the credibility of news sources (source-based fake news detection).

\section{Data Collection} \label{sec:data-collection}

The POLygraph: Polish Fake News Dataset was collected entirely from the Internet. The research team designed a mechanism using two methods: API data access and web scraping. For Twitter (nowadays X), we utilized the Twitter API\footnote{\url{https://developer.twitter.com/en/docs/twitter-api}}, which provided a powerful set of tools for Academic Researchers\footnote{\url{http://web.archive.org/web/20230212021429/https://developer.twitter.com/en/products/twitter-api/academic-research}} at the time. This allowed us to access archived data without putting additional strain on web services. The functions and methods provided in the API allowed us to search and filter the entire available content of Twitter freely, going all the way back to the first published tweet in 2006\footnote{\url{https://twitter.com/jack/status/20}}. We downloaded tweets from 2021-01-01 to 2022-04-30 to match the timeframe of other data sources. Twitter API provided the ability to search the entire archive and download up to 10 million tweets.
For websites, a custom scraper was employed to extract and save only the relevant content.

\subsection{Sources, Contents, and Authors}

The database of 5,000 sources was prepared by scraping a list of 1,300 starter websites. The scraper then visited at least 25 documents from each page and extracted subsequent links to external documents. Then, it repeated the process of searching and archiving documents. 
The XPath expression used to extract links from documents\footnote{\texttt{response.xpath("//body//a[not(starts- with(@href,'mailto:'))][not(starts-with (@href,'tel:'))]/@href").getall()}} provided the ability to retrieve all links whose \texttt{href} attribute does not start with \texttt{mailto:} or \texttt{tel:} and then return them as a list. In the next step, this list was iterated, and each address was passed to the parser, which added the address to the internal queue.
The scraped pages were archived as HTML files with linked materials in a structure consistent with the command \texttt{wget -H -k -r -l 1 url}.
The downloaded HTML files were automatically anonymized and then compressed into a ZIP archive, taking as the name documents a 128-bit hash function calculated based on the URL of the archived document.

\subsection{Tagged Press Articles} \label{sec:press-articles-database}

The aim of this stage of data collection was to create a database of about 3,000 tagged press articles. For this purpose, we queried Twitter to search for tweets whose content would be related to commenting on the truthfulness of the information, particularly expressing the opinion that some content constitutes fake news. We expected that entries of this type would contain references to newspaper articles and other sites that would be interesting to annotate for potential false information.
To obtain the URLs we were interested in, we used access to the Twitter API. We performed two variants of this search, differing in the query used and the time frame, resulting in two sets of entries:
\begin{itemize}
    \item V2 dataset -- a query focused on finding tweets where the author directly expresses their opinion on whether something is fake or not; uses phrases like “it wasn't fake” and “it was fake” in Polish and English\footnote{\texttt{(lang:pl (fejk OR fake OR fakenews OR "to nie był fake" OR "to był fake" OR "to nie był fejk" OR "to był fejk")) OR (lang:en (fejk OR "to nie był fake" OR "to był fake" OR "to nie był fejk" OR "to był fejk"))}} (1--29 April 2022; 574,545 obtained entries).
    \item V3 dataset -- a query like in V2, but extended with terms for debunking or verifying information, e.g., “verified”, “correction”, “where is this info from”\footnote{\texttt{(lang:pl (fake OR fakenews OR "fake news" OR factcheck )) OR ("to byl fejk" OR "to byl fake" OR "to nie byl fejk" OR "to nie byl fake" OR fejk OR "fejk-njus" OR dementi OR zweryfikowane OR "zrodlo potwierdzone" OR sprostowanie OR sprostowane OR "skad to info" OR "skad ta informacja" OR "przepraszam za podanie")}} (1 January 2010--31 July 2022; 3,580,901 obtained entries). 
\end{itemize}

In total, we collected 4,155,446 tweets.
Using a script to extract URLs from text, we obtained 339,259 URLs from this set.
% \texttt{select\_urls.py}

% \texttt{urls2jsonl.py}
The list of URLs was processed with another script, which uses Mercury Parser\footnote{\url{https://hub.docker.com/r/wangqiru/mercury-parser-api}}, \texttt{html2text}\footnote{\url{https://github.com/Alir3z4/html2text}}, and BeautifulSoup\footnote{\url{https://www.crummy.com/software/BeautifulSoup}} to extract text from the website located at the given URL. During the script execution, the following are rejected:
\begin{itemize}
    \item pages for which Mercury Parser found no text,
    \item pages for which the HTML returned by Mercury Parser was empty,
    \item pages that failed to convert HTML to text with either \texttt{html2text} or BeautifulSoup,
    \item pages whose language, detected based on the text using the \texttt{langdetect5} library, was other than Polish,
    \item pages for which \texttt{langdetect5} was unable to detect the language,
    \item repeated pages.
\end{itemize}

As a result, we received 63,776 examples in the JSON format supported by the Doccano \citep{doccano} annotation tool.

% \texttt{jsonl2pngs.py}
To give annotators access to a website preview, we created a spider (web crawler) that takes screenshots of the pages referenced by the URLs in the list and saves them to PNG files. The script uses the Scrapy\footnote{\url{https://scrapy.org}} framework and the \texttt{splash}\footnote{\url{https://splash.readthedocs.io}} library.
%
% \texttt{filter\_dan\_by\_images.py}
Then, using another script, we filtered the obtained examples in JSON format, discarding those for which it was impossible to take a screenshot of the page. Ultimately, we received 7,242 examples in JSON format (for Doccano), divided into 19 packages of 400 examples each (the last package was incomplete).
In this way, a collection of articles was prepared for detailed tagging. Each example in the collection was designated for annotation by at least three independent annotators. The annotation was carried out using the Doccano platform, as described in Section~\ref{sec:annotation-methodology-dan}.

\subsection{Tweets Expressing Opinions about Press Articles} \label{sec:opinion-tweets-database}

The starting point for obtaining a database of tweets expressing opinions about press articles was the dataset of 4,155,466 tweets described in Section~\ref{sec:press-articles-database}.
The subsequent processing stage was to extract external URLs of websites in Polish from this set of tweets. We wanted the resulting list of URLs to be both representative and diverse. To achieve this, we only considered one entry from each author and discarded URLs obtained through URL shorteners because they were likely redirects to other URLs in the set. Of the 4,155,446 tweets we rejected:
\begin{itemize}
    \item 3,249,033 tweets that did not refer to any external URL,
    \item 466,002 tweets in a language other than Polish,
    \item 197,208 tweets whose author was repeated,
    \item 63,885 tweets that contained more than one link to an external URL, and it was not possible to clearly indicate which of them they directly referred to,
    \item 46,665 tweets containing a URL that was most likely obtained using a shortener,
    \item 38,720 tweets containing the URL of a fact-checking website,
    \item 18,999 tweets containing an invalid URL.
\end{itemize}
74,934 examples left.

We wrote a Python spider called \texttt{tsv2pngs} using the Scrapy framework and the \texttt{splash} library. For each example from the source \texttt{data.tsv} file, the spider takes a screenshot of the tweet and a screenshot of the page the tweet refers to, combines them and saves the result as a PNG file. To access tweet content more easily, we used the Nitter service, which is a free, open-source front-end Twitter mirror. Before combining the screenshots, we scale them as needed to ensure the resulting PNG file is readable for annotators. Screenshots with aspect ratios (picture height to width ratio) greater than 8:1 are rejected. As a result, we obtained 22,206 PNG images of page screenshots.
% \texttt{tsv2jsonl.py}
A script that transforms data from TSV to JSONL files allowed us to obtain 74,934 examples in JSONL format.
% \texttt{filter\_jsonl.py}
An additional script utilizes the \texttt{urllib} library to filter out specific examples from the input file, including those without corresponding PNG screenshots, those with repeated website domains, and those that are part of a user-provided list. In our case, we supplied a list of examples annotated as part of the pilot annotation.
Ultimately, we ended up with 8,108 examples divided into three packages, which constituted data for three “fake-they-say” annotation tasks on the Doccano platform, described in Section~\ref{sec:annotation-methodology-fts}.

\section{Data Annotation}

\subsection{\emph{Fake-or-not} Annotation Methodology} \label{sec:annotation-methodology-dan}

The starting point for creating a set of questions for annotators in the discussed POLygraph dataset was the annotation scheme used in research on fake news in Japanese media by \citet{murayama-etal-2022-annotation}. The cited researchers proposed an annotation scheme that includes seven types of information: 1) the factuality of the news, 2) the disseminator's intention, 3) the target of the news, 4) the sender's attitude towards the recipient, 5) purpose of the news, 6) degree of social harmfulness of the news, 7) the type of harm that the news can cause.

The above set of questions is multidimensional, as it allows for considering a more comprehensive range of information than just the factuality aspect of the news. However, our catalog of questions expands beyond the above data. Although it is dominated by a text-centric approach, the questions are also aimed, among others, at determining the annotator's attitude towards the content, which helps recognize their bias and emotions evoked by the text. The detailed list of all 19 questions used in this annotation and related statistics are presented in Appendix~\ref{sec:appendix}.

The annotation was performed on the Doccano platform by a total of 161 annotators. The annotators in this task were experts and students of political sciences and journalism (see Section~\ref{sec:ethics}). All annotators underwent detailed training, including special case analysis. The total number of annotated news articles was 7,006, including 6,339 articles annotated by at least two independent annotators. The level of agreement between annotators was estimated by calculating Fleiss' kappa and varied depending on the question.

It is worth noting that our questionnaire contained many subjective and ambiguous questions because we wanted to investigate fake news in depth. Therefore, we do not expect perfect agreement among human annotators, especially when dealing with ambiguous or controversial cases. The nuanced nature of fake news detection further contributes to this expectation. The agreement scores reported by other studies on similar tasks take values around $0.3 \sim 0.4$. For instance, the RumourEval 2019 shared task achieved a Fleiss’ kappa of 0.39 for veracity annotation and 0.35 for stance annotation \cite{gorrell-etal-2019-semeval}. Thus, we believe that kappa scores within these limits would confirm the dataset’s usefulness for the purpose for which it was built.

\subsection{Gonito.net Platform}

We used the Gonito.net \citep{gonito2016} platform with the GEval \citep{gralinski-etal-2019-geval} evaluation tool to store and manage training, validation and testing data and evaluate the models used in the project. Gonito.net is an open-source platform for comparing and evaluating machine learning models, enabling reproducibility of experiments. On the Gonito.net platform, individual machine-learning tasks are organized as so-called challenges. A challenge is a set of training, validation and test data stored in a Git repository, associated with a set of evaluation metrics. Solutions to individual challenges can be put on the platform (in the form of model prediction results on a test set), which are automatically assessed using the GEval tool according to metrics related to the challenge. We have prepared two challenges for the project: \emph{fake-or-not} and \emph{fake-they-say}.

\subsection{\emph{Fake-or-not} Challenge} \label{sec:fake-or-not}

\begin{figure}[t]
    \includegraphics[width=\columnwidth]{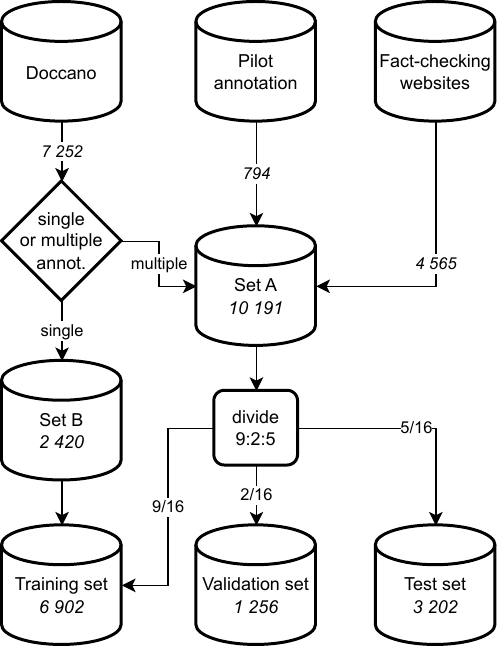}
    \caption{Data acquisition and processing workflow for \emph{fake-or-not} challenge.}
    \label{fig:data-flow}
\end{figure}

The \emph{fake-or-not} challenge is to create a model that will determine whether the article underneath it is fake news or not, based on the URL. The data for the challenge comes from three sources, detailed descriptions of which are provided below in the appropriate subsections: pilot annotation (Section~\ref{sec:fon-pilot}), annotation tasks on the Doccano annotation platform (Section~\ref{sec:fon-massive}), and fact-checking websites (Section~\ref{sec:fon-factcheck}).
Based on these three sources, a dataset (set~A) was created containing 10,191 records -- pairs: URL, label \texttt{1} (fake news) or \texttt{0} (not fake news).
Set~A was split in the proportions 9:2:5 into a training set (4,482 records), validation set (1,256) and test set (3,202). The split was made deterministically -- based on the last hexadecimal digit of the MD5 hash function value for the URL.
Additionally, set~B was obtained from annotation tasks on the Doccano platform, containing 2,420 analogous records (pairs: URL, label \texttt{1}/\texttt{0}). Set~B has been fully included in the training set.
To sum up, we have a total of 6,902 records in the training set, 1,256 records in the validation set and 3,202 records in the test set.
Out of all 11,360 records, there were 4,350 records marked with label~\texttt{1} and 7,116 records marked with label~\texttt{0}.
The diagram in Figure~\ref{fig:data-flow} summarizes the whole process.

\subsubsection{Pilot Annotation Task} \label{sec:fon-pilot}

As part of the annotation pilot, we prepared a set of 998 URLs of press articles. The method of collecting data is described in Section~\ref{sec:data-collection}. Each article was annotated by two independent annotators with one of three labels: “fake news”, “truth”, and “unknown”. The inter-annotator agreement measured by Cohen’s kappa was $0.421$. Then, URLs marked as “fake news” or “truth” by at least one annotator and for which both annotators’ annotations did not conflict were labeled \texttt{1} and \texttt{0}, respectively.
This way, we obtained 794 records (97 with label~\texttt{1} and 697 with label~\texttt{0}), which were added to set~A.

\subsubsection{Massive Annotation Task} \label{sec:fon-massive}

Annotation of the tasks described in Section~\ref{sec:annotation-methodology-dan} consisted of answering 19 detailed questions about the text. We only used the answers to question 12 to prepare data for the fake-or-not challenge (“In your opinion, does the text contain false information?”). Annotated examples for which the annotator chose the answer “yes” or “no” were selected (the answer “not subject to assessment” was omitted). Replies have been grouped by the related URLs. If the majority of the annotations for a given URL were “yes”, then a record consisting of the URL and label~\texttt{1} was added to the dataset, whereas if the majority of the annotations for the given URL were “no”, a record consisting of the URL address and label~\texttt{1}. The URL was omitted in case of an equal number of “yes” and “no” annotations.

Additionally, the obtained records were divided into two sets, depending on how many majority annotations there were for a given URL. If only one annotator indicated the majority answer (this also means that no annotator indicated the minority answer), the record was put in set~B. Otherwise, i.e., if at least two annotators indicated the majority answer, the record was put in set~A.
This way, we obtained 7,161 records, with 4,751 records in set~A and 2,410 in set~B. The label distribution is shown in Table~\ref{tab:doccano-label-distribution}.

\begin{table}
  \centering
    \begin{tabular}{lrrr}
      \hline
      \textbf{Set} & \textbf{Label~\texttt{1}} & \textbf{Label~\texttt{0}} & \textbf{Total} \\
      \hline
      A & 354 & 4,397 & 4,751 \\
      B & 1,179 & 1,231 & 2,410 \\
      \hline
      Total & 1,533 & 5,628 & 7,161 \\
      \hline
    \end{tabular}
    \caption{Label distribution}
    \label{tab:doccano-label-distribution}
\end{table}

\subsubsection{Data from Fact-checking Websites} \label{sec:fon-factcheck}

Opinions from fact-checking websites (476 opinions from \url{fakehunter.pap.pl}, 2,125 opinions from \url{demagog.org.pl}, and 2,637 reviews from \url{afp.com}) were used as another source of data.
If the opinion was expressed as “fake news”, “false”, “manipulation”, etc., a record consisting of the appropriate URL address and label~\texttt{1} was added to the dataset. If the opinion was expressed as “true”, the appropriate record was tagged with~\texttt{0}.
This way, we obtained 4,924 records (3,784 with label~\texttt{1} and 1,140 with label~\texttt{0}), which were added to set~A.

\subsection{\emph{Fake-they-say} Annotation} \label{sec:fake-they-say}

\subsubsection{Annotation Methodology} \label{sec:annotation-methodology-fts}

The “fake-they-say” annotation task was developed to assess the degree of the tweet author's belief in the (un)truthfulness of the information they commented on. The annotators received access to the content of 1) the tweet being rated, 2) the entire discussion regarding the news, and 3) the news itself. The task was to read the content of the comment on a specific piece of news and/or the entire accompanying discussion and then select one of the following six labels defining the tweet author's attitude towards the content of the article:
\begin{itemize}
    \item \emph{hard-claim-fake} (the author of the tweet claims that the news they are commenting on is false),
    \item \emph{hard-claim-not-fake} (the author of the tweet claims that the news they are commenting on is true),
    \item \emph{no-claim} (it is impossible to determine what the author of the tweet thinks, or the comment does not refer to the issue of (un)truthfulness of the news),
    \item \emph{sarcasm} (the author of the tweet is ironizing, expressing themselves sarcastically),
    \item \emph{soft-claim-fake} (the author of the tweet probably believes that the news they are commenting on is false),
    \item \emph{soft-claim-not-fake} (the author of the tweet probably does not think the news they are commenting on is false).
\end{itemize}

The annotators in this task were experts and students of political sciences and journalism. All annotators underwent detailed training.
There were 48 annotators, and they annotated 4,356 press articles in total, including 3,235 articles annotated by at least two independent annotators. The level of agreement between annotators was estimated by calculating Fleiss' kappa as $\kappa = 0.4343$.

\subsubsection{Challenge Description}

The \emph{fake-they-say} challenge is to create a model that, based on the tweet’s text and the URL, will determine what the tweet’s author thinks about the article located at the given URL.
The data for the challenge comes from two sources (detailed descriptions provided below in the relevant subsections): pilot annotation and annotation tasks on the Doccano annotation platform.
These two sources created a dataset containing 5,082 records, consisting of the following fields:
\begin{itemize}
    \item label: one of the 6 labels described in Section~\ref{sec:opinion-tweets-database} (\emph{hard-claim-fake}, \emph{hard-claim-not-fake}, \emph{no-claim}, \emph{sarcasm}, \emph{soft-claim-fake}, \emph{soft-claim-not-fake}),
    \item tweet text,
    \item tweet URL,
    \item URL address of the commented article,
    \item PNG image consisting of a screenshot of the tweet and a screenshot of the commented article.
\end{itemize}

The dataset was split in the proportions 13:1:2 into the training set (4,040 records), validation set (316 records) and test set (726 records). The split was made deterministically -- based on the last hexadecimal digit of the MD5 hash function value for the URL. In total, we obtained 806 \emph{hard-claim-fake} records, 102 \emph{hard-claim-not-fake} records, 1,254 \emph{no-claim} records, 44 \emph{sarcasm} records, 421 \emph{soft-claim-fake} records and 166 \emph{soft-claim-not-fake} records.

\subsubsection{Pilot Annotation Data}

As part of the annotation pilot, we prepared a collection of 1,000 tweets referring to various URL addresses. The method of collecting data is described in Section~\ref{sec:data-collection}. Each tweet was annotated by 4 independent annotators with one of the 6 labels described in Section~\ref{sec:opinion-tweets-database} (\emph{hard-claim-fake}, \emph{hard-claim-not-fake}, \emph{no-claim}, \emph{sarcasm}, \emph{soft-claim-fake}, \emph{soft-claim-not-fake}). Then, the annotations for each tweet were aggregated according to the following algorithm:

\begin{enumerate}
    \item If all annotators have chosen the same label, assign that label.
    \item Otherwise:
    \begin{itemize}
        \item if any annotators have chosen the label \emph{*-claim-fake} and no annotators have chosen the label \emph{*-claim-not-fake}, assign the label \emph{soft-claim-fake},
        \item if any annotators have chosen the label \emph{*-claim-not-fake} and no annotators have chosen the label \emph{*-claim-fake}, assign the label \emph{soft-claim-not-fake}.
    \end{itemize}
    \item In other cases, assign the label \emph{no-claim}.
\end{enumerate}

This way, we obtained 1,000 records.

\subsubsection{Data from Annotation Tasks on the Doccano Platform}

The method of collecting data for annotation tasks is described in Section~\ref{sec:data-collection}.
Annotation in these tasks consisted of selecting one of the 6 labels described in Section~\ref{sec:opinion-tweets-database} (\emph{hard-claim-fake}, \emph{hard-claim-not-fake}, \emph{no-claim}, \emph{sarcasm}, \emph{soft-claim-fake}, \emph{soft-claim-not-fake}) based on the text of the tweet and the content of the website to which the tweet concerned. Then, the annotations for each tweet were aggregated according to the same algorithm as in the case of the pilot annotation.
This way, we obtained 4,082 records.

\section{Anonymization/Privatization} \label{sec:anonymization}

Privatization is an important step in the process of constructing any language resource that combines news and social media text.
It requires thoughtful planning with regard to the categories of personal identifiable data that should or should not be anonymized.
On the one hand, the names of public figures and coarse-grained descriptions of geographical locations of events are not considered private.
Thus, they should not be anonymized in the corpus.
On the other hand, the names of private citizens, their home addresses or any other personal identifiable information should be removed.
To solve this problem, we developed a privatization tool that consists of three modules: 1) named entity recognizer, 2) alphanumeric expression classifier, and 3) privacy checker.

The named entity recognizer follows Transformer architecture \cite{vaswani17} and utilizes a pre-trained language model \cite{devlin-etal-2019-bert}.
It is based on the HerBERT model\footnote{\url{https://huggingface.co/allegro/herbert-base-cased} }\cite{mroczkowski-etal-2021-herbert} with a token classification head attached.
The alphanumeric expression classifier is responsible for detecting potentially private phrases with strict definitions that can be described using regular expressions.
The categories of expressions identified by this module are summarised in Table~\ref{tab:alphanumcls}.
The privacy checker considers all expressions detected by the named entity recognizer and the alphanumeric expression classifier to be private by default.
It makes public only the names that appear in an index of public figures built on the basis of DBpedia \cite{lehmann15} entries that belong to
the \url{<https://dbpedia.org/ontology/Person>} class in the DBpedia ontology, denoted by the Polish or English language code.

\begin{table}
  \centering
    \begin{tabular}{ll}
        \hline
        \textbf{Category} & \textbf{Description} \\
        \hline
         url & uniform resource locator \\
         email & e-mail address \\
         cardnumber & credit/debit card number\\
         zipcode & postal code\\
         username & username in social media\\
         nip & tax ID\\
         passport & passport number \\
         idcard & identity card number\\
         crypto & crypto wallet address \\
         macaddr & MAC address \\
         accountnumber & bank account number \\
         address & physical address \\
         phone & phone number\\
         \hline
    \end{tabular}
    \caption{Categories of data detected by the alphanumeric expression classifier.}
    \label{tab:alphanumcls}
\end{table}

\section{Dataset Summary and Discussion}

The POLygraph Polish fake news dataset consists of two parts: \emph{fake-or-not} and \emph{fake-they-say}, which are detailed in Sections \ref{sec:fake-or-not} and \ref{sec:fake-they-say}.Together, they form a new dataset for detecting fake news in Polish. Unlike existing datasets, this dataset is not solely or predominantly based on a binary true-false classification but draws on various approaches proposed in source literature.
The overview of the dataset is shown in Table~\ref{tab:overview}.

\begin{table}
  \centering
    \begin{tabular}{lrrr}
      \hline
      \textbf{Set} & \textbf{fake-or-not} & \textbf{fake-they-say} \\
      \hline
      Training set & 6,902 & 4,040 \\
      Validation set & 1,256 & 316 \\
      Test set & 3,202 & 726 \\
      \hline
      Total & 11,360 & 5,082 \\
      \hline
    \end{tabular}
    \caption{The POLygraph dataset summary}
    \label{tab:overview}
\end{table}

This approach results in collecting a range of data typically utilized in news-content-based, knowledge-based, and user-information-based fake news detection methods. Although the POLygraph dataset has not yet been used in real-world scenarios, it was developed for a project aimed at verifying information sources and detecting fake news. Further exploration of the collected data by an interdisciplinary team of researchers will foster fake news detection and provide institutions and scholars with a more comprehensive range of data than previous fake news datasets. The envisioned use case involves building tools that detect false information and mark such information in search engines, potentially tested by monitoring social media messages over some time.

Additionally, adapting the POLygraph dataset for other languages should not pose a significant problem. The dataset itself is based on solutions proposed for other languages, often very different from one another, such as English and Japanese. This universality strengthens the argument that the core concept can be applied across various languages and cultural settings. Some proposed solutions might require modifications depending on the specific language, but the core strength remains – the applicability across diverse contexts. The presented annotation scheme will hopefully serve as a stimulus for implementing an analogous detection model for other languages.

\section{Acknowledgments}

This research was conducted as part of a project “From Urban Legend to Fake News. A Global Detector of Contemporary Falsehoods”, funded by the Polish National Centre for Research and Development, grant number INFOSTRATEG-I/0045/2021.

\section{Limitations}

This study acknowledges the inherent challenges in building a comprehensive fake news detection system. The dataset, while extensive, might not capture every form of misinformation online, limiting the generalizability of the findings. Additionally, the use of human annotation introduces subjectivity, as annotators may have differing definitions of what “fake news” is. Including subjective and ambiguous questions to explore fake news in depth can lead to disagreements, especially in borderline cases. However, perfect agreement is not expected in such nuanced tasks – similar projects report agreement scores around $0.3 \sim 0.4$ \cite{gorrell-etal-2019-semeval}, which is deemed acceptable here.

The complexity of fake news detection is reflected in the multidimensional annotation scheme employed. This paper focuses on data collection and annotation, with the evaluation of the dataset’s efficacy in machine learning tasks planned for a future stage. Similarly, the description and evaluation of a potential fake news recognition tool using this dataset are beyond the scope of this article.

Furthermore, the study primarily focuses on the Polish language, limiting its direct applicability to other languages and cultures. The ever-evolving nature of fake news tactics also necessitates continuous updates to the dataset and any future detection tool to maintain effectiveness.

Despite these limitations, this study offers valuable insights into fake news detection and lays a robust foundation for future research in this area.

\section{Ethics Statement} \label{sec:ethics}

The human annotators were recruited from a pool of student volunteers who expressed interest in participating in the project. They were informed about the project's purpose, methods, and expected outcomes, and they gave their consent before starting the annotation task. They were given clear instructions and guidelines for the annotation task and received feedback and support whenever needed. They were free to withdraw from the project at any time without any negative consequences. The annotators did not receive payment for participating in the project, as they agreed to volunteer their time and effort for scientific research. The authors have the right to use the data presented in the paper, and they ensured that the data was anonymized and privatized to protect the privacy and confidentiality of the individuals and entities involved.

\bibliography{anthology,bibliography}

\appendix

\section{Appendix: Annotation Questions}
\label{sec:appendix}

\renewcommand{\labelenumi}{Q\arabic{enumi}:}
\renewcommand{\labelenumii}{\alph{enumii})}

\begin{enumerate}
\item Specify the type of text.

\begin{enumerate}
\item article on a news website

\item social media post

\item blog post

\item other
\end{enumerate}

\item Define the subject matter of the text.

\begin{enumerate}
\item politics

\item society

\item medicine

\item military

\item economy

\item entertainment

\item education

\item science and technology

\item tourism

\item culture

\item sports

\item business

\item crime

\item safety

\item religion

\item other

\end{enumerate}
\item What is your attitude to the text?

\begin{enumerate}
\item I agree with the text.

\item I do not agree with the text.

\item I have a neutral attitude to the text.

\end{enumerate}
\item What emotions does the text evoke in you?

\begin{enumerate}
\item positive

\item negative

\item The text does not evoke emotions in me.

\end{enumerate}
\item What content dominates in the text?

\begin{enumerate}
\item facts

\item opinions

\item both

\end{enumerate}
\item Is the text persuasive?

\begin{enumerate}
\item yes

\item no

\item I don't know

\end{enumerate}
\item What do you think is the purpose of the news?

\begin{enumerate}
\item information - the text is purely informative, it reports and describes events

\item disinformation - the author deliberately provides false information in order to obtain some benefits (e.g. political or financial)

\item propaganda - the text is persuasive and affects the emotions, attitudes, opinions and/or actions of the target audience for ideological, religious and other purposes

\item partisan promotion of political views - the text presents information in a biased way from the perspective of a specific political party or political ideology

\item entertainment (satire / parody) - the purpose of the text is to provide the target with entertainment and / or criticism of individuals or groups

\item other

\end{enumerate}
\item Who do you think is the potential target of the news?

\begin{enumerate}
\item recipient of general news from news websites

\item recipient of entertainment

\item supporter of a specific political party

\item supporter of a specific socio-political ideology

\end{enumerate}
\item Does the author/disseminator believe that the news they are writing about is true?

\begin{enumerate}
\item yes, the author openly expresses the belief that they agree with what they are disseminating

\item yes; however, the author expresses doubts about the veracity of the news

\item no, the author openly denies the veracity of the news

\item no comments are made by the author

\end{enumerate}
\item Does the author refer to the sources of the cited information?

\begin{enumerate}
\item yes

\item no

\item sometimes / not always

\end{enumerate}
\item What narrative style is the main basis of the news?

\begin{enumerate}
\item conflict (often specific to political events, centered around disagreement, division, difference or rivalry)

\item responsibility (assigning responsibility for the cause/effect of the presented problem to specific persons/institutions etc.)

\item morality (related to the moralizing tendencies of the media; it most often refers to condemnation or other forms of moral evaluation of the presented events)

\item human story (personalization which introduces emotional elements, the main character is most often the victim of a tragic event or crisis; greater importance is attached to the individual affected by the event than its global consequences)

\item consequences (related to a broader context and impact on various areas of social life)

\end{enumerate}
\item In your opinion, does the text contain false information?

\begin{enumerate}
\item yes

\item no

\item not subject to assessment (the text contains only the author's opinion)

\end{enumerate}
\item What kind of false information is contained in the text?

\begin{enumerate}
\item fake news - false information has been included in the article intentionally and it is possible to verify it (without referring to external sources!)

\item rumor - the author refers to unconfirmed information (e.g. rumors)

\item satire - the author cites false information that is humorous, ironic, mocking; it is not intended to mislead the reader

\item clickbait - the title attracts attention, but does not reflect the content of the news

\end{enumerate}
\item Where is the false information located in the text?

\begin{enumerate}
\item in the title/headline

\item in one fragment

\item false information is repeated in several fragments of the text

\item in the image

\item the whole text is false

\end{enumerate}
\item How much of the text must be read in order to realize that it contains false information?

\begin{enumerate}
\item headline / title

\item the title and part of the text

\item the entire text

\end{enumerate}
\item If the news contains false information, do you think the author of the text knows that they are disseminating false information?

\begin{enumerate}
\item They know it.

\item They probably know it.

\item They don't know it.

\item They definitely don't know it.

\end{enumerate}
\item Have you come across the false information contained in the text before?

\begin{enumerate}
\item Yes.

\item No.

\end{enumerate}
\item How socially harmful is the false information contained in the text?

\begin{enumerate}
\item 0 (harmless)

\item 1 (slight harm, e.g. lack of understanding of certain events)

\item 2 (moderately harmless, e.g. causing confusion and anxiety)

\item 3 (moderately harmful, e.g. leading to conspiracy theories)

\item 4 (relatively harmful, e.g. damage to the reputation of people and institutions, prejudice against a nation, race etc.)

\item 5 (very harmful, e.g. health and life hazard)

\end{enumerate}
\item What kind of threat may be posed by the false information?

\begin{enumerate}
\item lack of understanding of political and social events

\item damage to the reputation of persons and institutions, undermining trust in persons and institutions

\item prejudice against nation, race, state

\item confusion and fear of society

\item the emergence of conspiracy theories

\item risk to health and life

\item none
\end{enumerate}
\end{enumerate}

\end{document}